\documentclass[final]{cvpr}

\usepackage{xspace}
\usepackage{times}
\usepackage{epsfig}
\usepackage{graphicx}
\usepackage{amsmath}
\usepackage{amssymb}
\usepackage{booktabs}
\usepackage{paralist}
\usepackage{textcase}
\usepackage{float}
\usepackage{subcaption}
\usepackage{stackengine}

\usepackage[pagebackref=true,breaklinks=true,colorlinks,bookmarks=false]{hyperref}

\def\cvprPaperID{11} 
\def\confYear{EarthVision 2021}

\begin{document}

\title{Towards Indirect Top-Down Road Transport Emissions Estimation}

\author{Ryan Mukherjee \qquad
 Derek Rollend \qquad
 Gordon Christie \qquad
 Armin Hadzic \\
 Sally Matson \qquad
 Anshu Saksena \qquad
 Marisa Hughes\\
Johns Hopkins University Applied Physics Laboratory\\
{\tt\small \{firstname\}.\{lastname\}@jhuapl.edu}
}


\twocolumn[{
\renewcommand\twocolumn[1][]{#1}
\maketitle
}]

\thispagestyle{empty}
\begin{abstract}
Road transportation is one of the largest sectors of greenhouse gas (GHG) emissions affecting climate change. Tackling climate change as a global community will require new capabilities to measure and inventory road transport emissions. However, the large scale and distributed nature of vehicle emissions make this sector especially challenging for existing inventory methods. In this work, we develop machine learning models that use satellite imagery to perform indirect top-down estimation of road transport emissions. Our initial experiments focus on the United States, where a bottom-up inventory was available for training our models. We achieved a mean absolute error (MAE) of 39.5 kg CO$_{2}$ of annual road transport emissions, calculated on a pixel-by-pixel (100~m$^{2}$) basis in Sentinel-2 imagery. We also discuss key model assumptions and challenges that need to be addressed to develop models capable of generalizing to global geography. We believe this work is the first published approach for automated indirect top-down estimation of road transport sector emissions using visual imagery and represents a critical step towards scalable, global, near-real-time road transportation emissions inventories that are measured both independently and objectively.
\end{abstract}

\section{Introduction}
\label{ch:introduction}

\begin{figure*}[t]
\includegraphics[width=\textwidth]{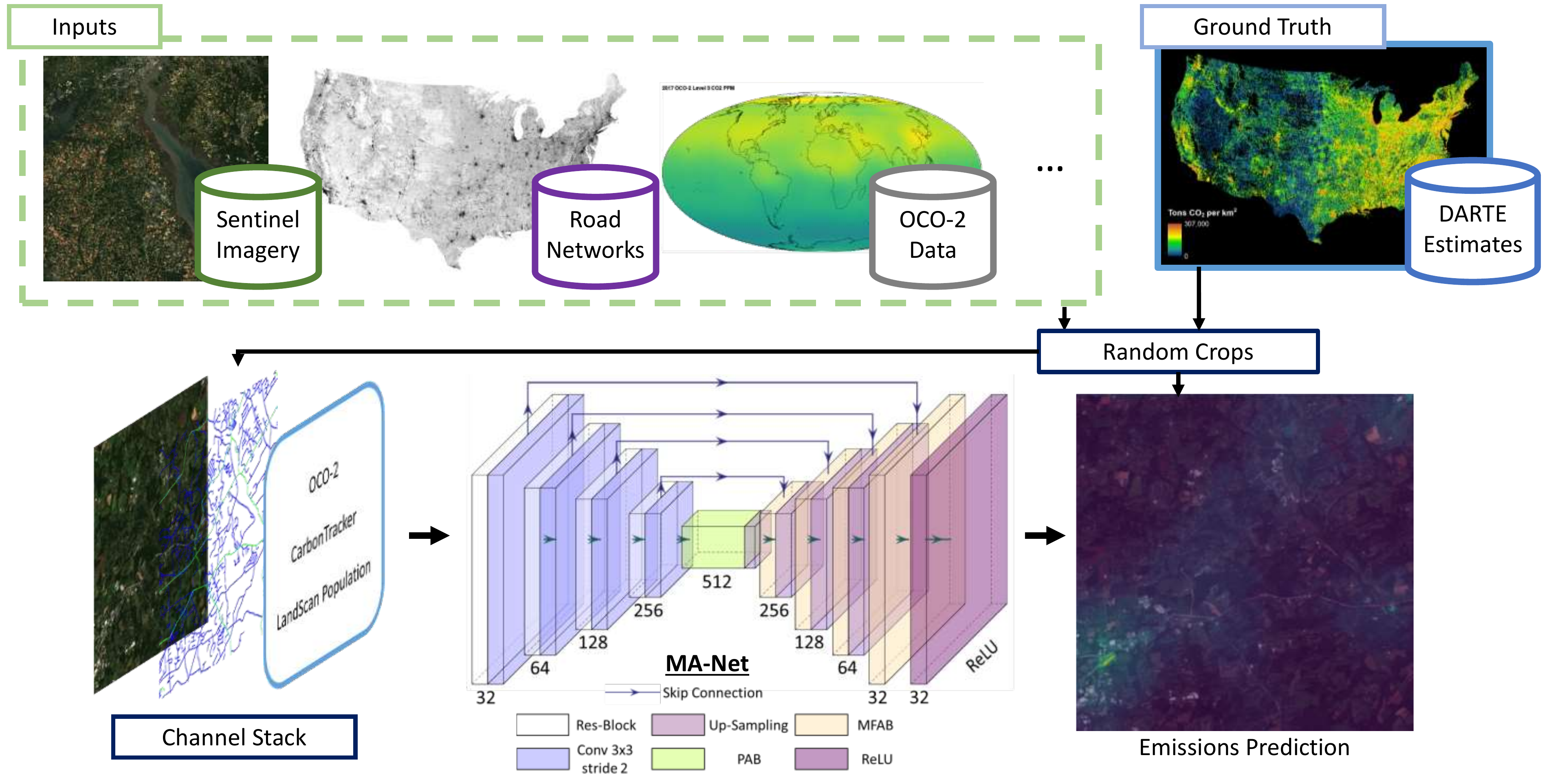}
\caption{An overview of our process for learning to regress road transport GHG emissions. Primary inputs to our model include Sentinel-2 satellite imagery (RGB bands only) and road networks from map data, but we also perform experiments using Orbiting Carbon Observatory-2 (OCO-2) data at 1$^{\circ}$ x 1$^{\circ}$ resolution ($\approx$100 x 100~km), NOAA CarbonTracker (CT) data, and LandScan population estimates. The ground truth for our experiments is derived from the Database of Road Transportation Emissions (DARTE) \cite{gately2015cities}, which provides annual bottom-up CO$_2$ estimates at a spatial resolution of 1~km$^2$. We crop and interpolate all inputs and ground truth to the resolution of the Sentinel-2 input imagery (10~m ground sample distance). We train a MA-Net that learns to regress per-pixel CO$_2$ values. This process enables us to improve the spatial resolution of DARTE estimates, and provide estimates for the specific point in time that the Sentinel-2 imagery was collected.}
\label{fig:teaser}
\end{figure*}

Transportation contributed 28\% of anthropogenic GHG emissions in the U.S. for 2018, higher than any other sector (electricity generation, the second highest, contributed 27\%)~\cite{epa2018, hockstad2018inventory}. The primary source of transportation sector emissions are vehicles, which account for 82\% of emissions.
Globally, road transport is also a significant contributor as it accounted for approximately 12\% of GHG emissions in 2016~\cite{climatewatch2020}. Addressing GHG emissions at a global scale will require reductions in emissions across many sectors, and perhaps most significantly to road transport.

Mitigating the impact of global warming will require transportation emissions to be taken into account. According to the goals established by the Paris Agreement, reducing transportation sector emissions is essential to maintaining global warming within 1.5$^{\circ}$ or 2$^{\circ}$C~\cite{sims2014transport}. Quantifying the distribution of on-road transportation emissions is vital to this reduction effort as inventories help identify trends, track mitigation efforts, and inform policy decisions.

Multiple efforts are developing detailed bottom-up on-road emission inventories for the U.S.~\cite{gately2015cities, gurney2020vulcan}. These projects are limited from expanding globally due to the reliance on vehicle traffic and road data that is not readily available on a global scale. EDGAR sought to improve on the scope of emissions data by providing a global inventory for transportation that uses road density as a proxy to downscale emissions geographically~\cite{janssens2017edgar}. However, some emission estimates for urban centers in EDGAR deviated from other bottom-up inventories \cite{gately2015cities} by 500\%, indicating that road density is not a sufficient proxy for global high-resolution inventories.

Our work seeks to build upon these previous on-road emissions inventory methods. Our approach leverages deep learning methods for indirect estimation of on-road emissions, at a global scale, with minimal region-specific tuning effort. Our models leverage satellite imagery as their primary input, enabling them to support increased spatial resolution and temporal frequency of on-road GHG estimates.

\section{Related Work}
\label{sec:related_work}

We provide an overview of the bottom-up measurement methodologies of road transport emissions inventories and emissions estimation using top-down remote sensing technologies. The UNFCCC requires Annex I countries to provide yearly bottom-up GHG emission inventories~\cite{unfccc2013} using standardized methodology from the IPCC~\cite{eggleston20062006}. These bottom-up inventories are generated using activity data (e.g., amount of fuel purchased) and emission factors (quantity of GHGs emitted per activity unit). Transportation emissions are calculated considering vehicle types and uses in order to provide robust and proper estimates.

EDGAR is a global-scale inventory that follows the IPCC methodology and thus acts as independent validation against self-reported figures, as well as a uniform comparison between countries~\cite{janssens2017edgar, muntean2018fossil}. EDGAR data is available on monthly timescales on a country and sectoral basis \cite{crippa2020high} or as a 1$^{\circ}$ gridded global annual product. For the road transportation sector, country or sub-country sectoral emissions are downscaled spatially according to road density.

Recent works, primarily focused the U.S., investigate on-road GHG emissions. Gately et al.~\cite{gately2015cities}, for example, use reported vehicular traffic (VHT) data combined with Census TIGER~\cite{marx1986tiger} road network information to estimate regional on-road emissions and disaggregate them among mapped road networks.

The ``Vulcan Product" is a national-scale, multi-sectoral, hourly inventory from 2010-2015 with a resolution of 1~km$^2$~\cite{gurney2020vulcan}. In this work, transportation emissions are based on EPA county-level on-road emissions estimates, which are further downscaled using data from the Federal Highway Administration. Gurney et al. also introduced HESTIA~\cite{gurney2012quantification}, a city-scale produce that uses additional city-specific data sources to improve inventory accuracy.

While these road transportation inventories use bottom-up methods, there are methods of measuring emissions that can be utilized to create inventories of the transportation sector. For example, spectral measurements are commonly used to directly measure the presence of GHGs by calculating atmospheric absorption features. This can be accomplished readily using ground-based sensors~\cite{peters2007atmospheric, toon2009total}, which are incorporated in some of the aforementioned bottom-up approaches. However, more recently, the availability of spectral-based satellite observations has been increasing~\cite{crisp2004orbiting, crisp2017orbit, buchwitz2015greenhouse}. While these systems have been used for direct GHG assessment as early as 1987~\cite{chedin2002annual}, only recently have technological advancements increased the appeal and accessibility of these systems. However, spectral-based satellite observations still remain fairly coarse in their spatial resolution and temporal frequency. We investigate using these observations as model inputs, but they may also be beneficial for validating finely-gridded sectoral estimates after some spatial aggregation.

Satellite imagery has also been used to estimate GHG emissions indirectly. Oda et al.~\cite{oda2011very, oda2018open} demonstrate that visual features, in this case night lights, can be used to disaggregate national emissions and create high-resolution emissions estimates grid. Couture et al.~\cite{couture2020emissions} present initial works towards using visible steam plumes to estimate the operational state of power plants, leading to plant emissions. Zheng et al.~\cite{zheng2020estimating} also show that convolutional neural networks (CNNs) can learn to estimate ground-level PM$_{2.5}$, a measure of mostly invisible air pollution, from 3-5m resolution visible-spectrum PlanetScope imagery \cite{planetscope2020spec}. With some additional metadata, including weather information, they show that their model can achieve MAE on the order of 20\%.  We  take inspiration from these recent efforts in order to create the first global, near-real-time road transportation emissions inventory.

\section{Methods}
\label{ch:method}

\subsection{Data}

\begin{figure}[t]
    \centering
    \includegraphics[width=0.47\textwidth]{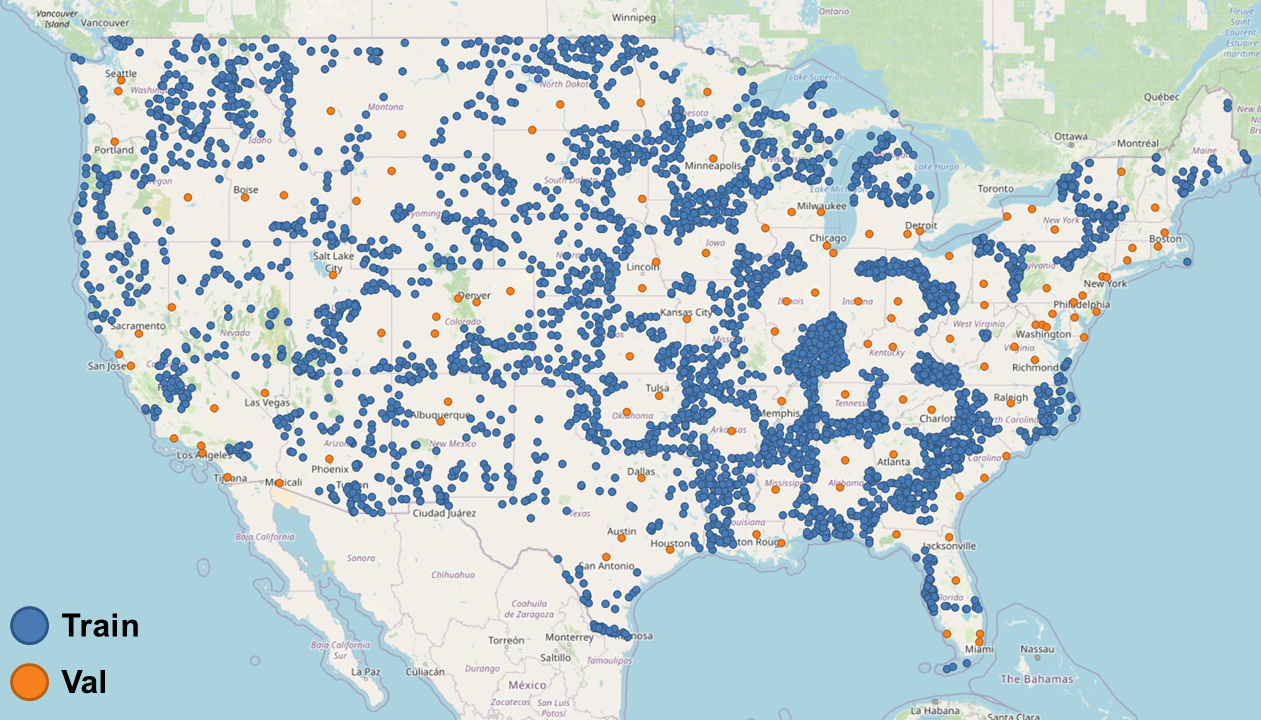}
    \caption{Plot showing city locations sampled from the conterminous United States (CONUS) for training (blue) and validation (orange).}
	\label{fig:trainval_locations}
\end{figure}

Data is perhaps the most critical component to developing a successful approach for top-down road transport emissions estimation. Vehicles are small, abundant, and frequently on the move, which makes them especially challenging to directly observe. Direct measurement of road transport emissions in a top-down manner would require substantial infrastructure and technological development to monitor and track all vehicles. For example, accomplishing this with Earth observation satellites would require new technological advancements addressing spatial resolution, night-time capability, and scalability for continuous monitoring. Currently, many relevant Earth observation systems are incapable of directly observing periods of peak vehicular activity (i.e., early morning or late afternoon commutes) because they operate in sun-synchronous orbits, such as NASA's Afternoon Constellation~\cite{schoeberl2002afternoon}. Additionally, many of these systems do not have sufficient spatial resolution to resolve vehicles. PlanetScope imagery~\cite{planetscope2020spec}, at 3-5~m resolution, seems to be near the limit of what can be used for monitoring vehicles~\cite{drouyer2020parking, chen2021spatial}.

To address these limitations, we investigate leveraging various alternative data sources that we hypothesize can be used to indirectly estimate road transport emissions. These data sources and how we use them in our models are described in the following subsections.

\begin{figure}[t]
    \centering
    \includegraphics[width=0.47\textwidth]{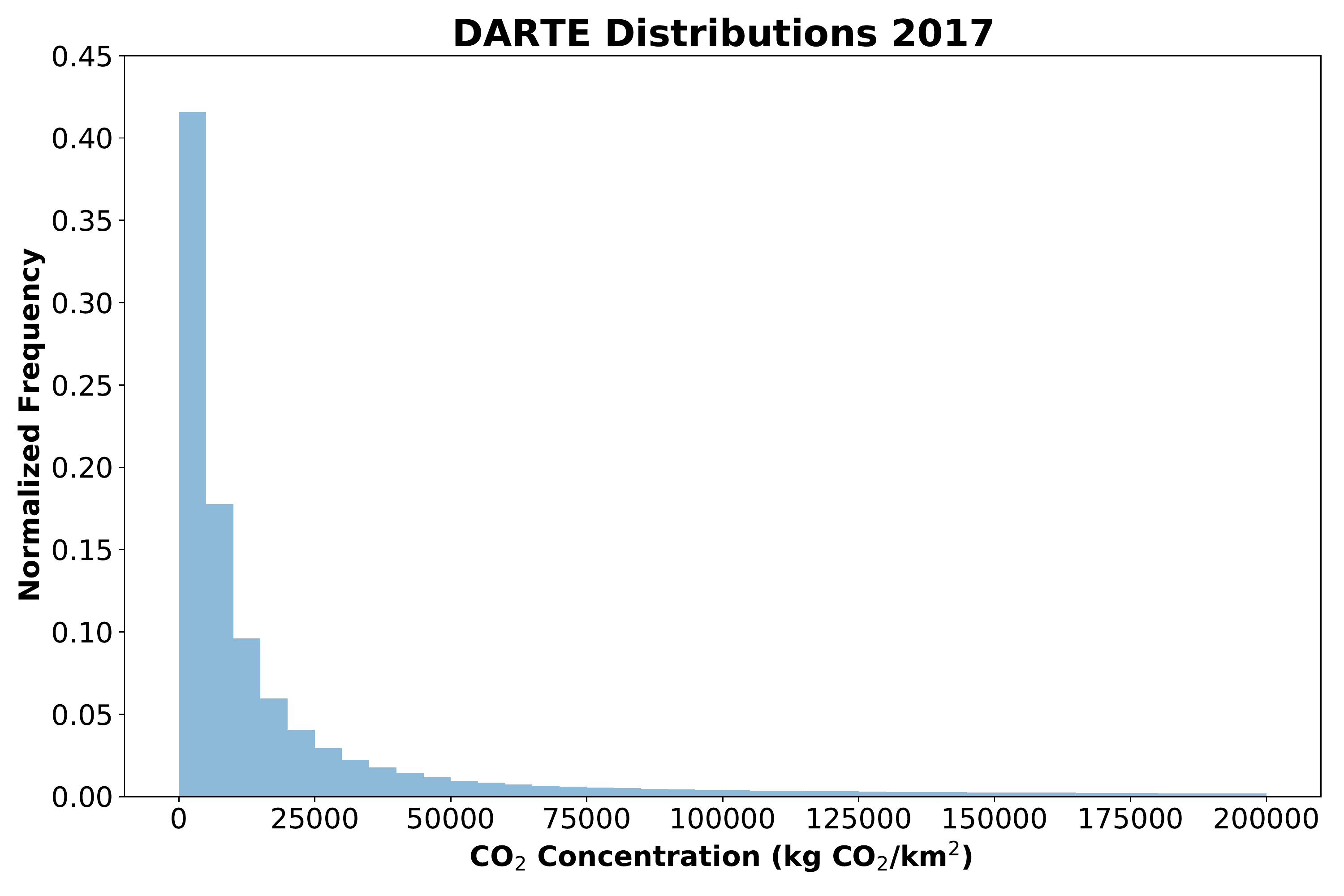}
    \caption{Histogram of DARTE road transport CO$_{2}$ emissions estimates for 2017}
    \label{fig:darte_distribution}
\end{figure}

\begin{figure*}[t]
    \centering
    \includegraphics[width=\textwidth,trim=0.3cm 2.6cm 0.3cm 2.6cm,clip]{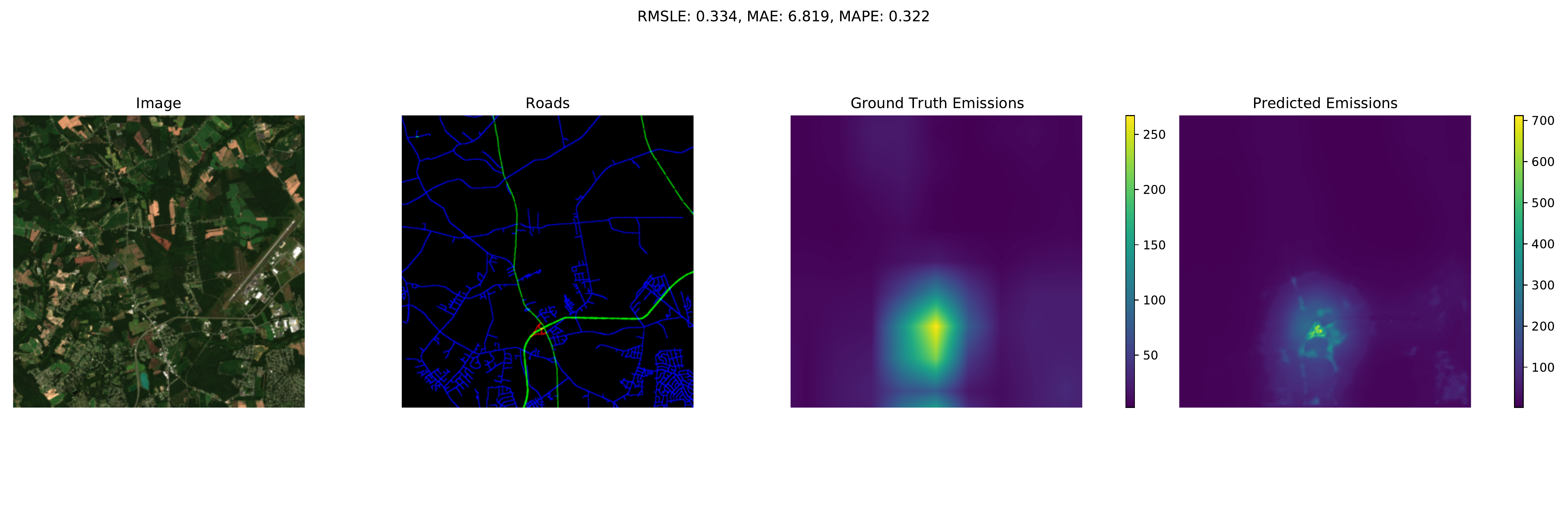}
    \caption{Sample results generated by the EN-B3 MA-Net model trained on Sentinel-2 and road data. Colorbars are in units of kg CO$_{2}$ per 100~m$^2$. Roads are colored according to their category: primary road and ramp (red), secondary (green), and local (blue).}
    \label{fig:sample_results}
\end{figure*}

\subsubsection{Road Transport Emissions}
Our models learn to regress road transportation CO$_{2}$ emissions using supervised training with the Database of Road Transportation Emissions (DARTE)~\cite{gately2015cities}. DARTE provides annual (1980-2017) bottom-up CO$_2$ estimates at a spatial resolution of 1~km$^2$ covering the conterminous United States. DARTE leverages the Highway Performance Monitoring System, which provides road segments with the following properties: annual average daily traffic (AADT), functional class, urban/rural context, and county. Road segment lengths were combined with AADT to calculate the annual vehicle miles travelled (VMT) per segment, which was partitioned into five vehicle types (passenger cars, passenger trucks, buses, single-unit trucks, and combination trucks). A calibrated fuel economy and VMT by vehicle type were used to calculate motor gasoline and diesel fuel consumption, which was smoothed and scaled so that intra-state values summed to the reported state totals. These motor gasoline and diesel fuel consumption estimates were then converted to CO$_{2}$ emissions on an annual basis. For this initial work we only use DARTE data for 2017, the distribution of which is shown in Figure~\ref{fig:darte_distribution}.

\subsubsection{Visual Imagery}
Visible-spectrum satellite imagery is the primary input for our models. We use Sentinel-2 Level-2A products~\cite{drusch2012sentinel, gatti2013sentinel} at 10 x 10~m (100~m$^2$) spatial resolution. Level-2A captures bottom-of-atmosphere reflectance and incorporates radiometric calibration and orthorectification corrections from previous product stages. Sentinel-2 collects 13 spectral bands with band centers ranging from approximately 443~nm to 2190~nm. We only use bands 4, 3, and 2 from each Sentinel-2 image, which roughly correspond to visible red, green, and blue channels, respectively. These bands are stacked to form a single 3-channel RGB image, which is also referred to as a true color composite.

As satellite imagery is the primary input to our model, we maintain the coordinate reference system (CRS) used by Sentinel-2 Level-2A products and map all auxiliary inputs and ground-truth measurements to this CRS. With the exception of road data, which is mapped to each swath, all other auxiliary inputs and ground-truth measurements are mapped to Sentinel-2 imagery on a tile-by-tile basis. In other words, during training we dynamically extract a more manageable subset of each Sentinel-2 swath, which we refer to as a tile, and simultaneously extract all corresponding auxiliary inputs and ground-truth measurements that fall within the bounds of this tile. To account for potential blank regions in each tile, a blank-region mask is created and used to zero out the corresponding regions from all auxiliary inputs and ground truth.

Swaths of Sentinel-2 imagery consist of 10800 x 10800 pixels covering approximately 11,664~km$^{2}$. A maximum of 5 Sentinel-2 swaths for each location in our dataset are temporally sampled from the summer (June 1st through September 30th) of 2017 to minimize the impact of seasonal changes and occlusion due to weather. To spatially tile the data, we use a systematic unaligned sampling process \cite{delmelle2009spatial}. The Sentinel-2 swaths are first equally divided into an 11x11 grid and then we use this grid as a starting point from which a random sub-tile-sized offset is calculated. The combination of a grid intersection and random offset determine the 1024 x 1024 pixel tile (approximately 105~km$^{2}$) of a Sentinel-2 swath that will be extracted for training. The advantage of using this systematic unaligned sampling during the training process is that it maintains an even sampling of tiles from each swath while also introducing a form of data augmentation by not continually sampling identical images.

Once Sentinel-2 tiles have been sampled, and before they are input into our models, we apply standard ImageNet normalization
and float conversion ($intensity\in[0,1]$). Auxiliary inputs, if any, are then concatenated as additional channels in these tile images\footnote{We do not mask clouds, but intend to explore it in future work.}.

\subsubsection{Roads}
Road network data is used as an input to our model for three reasons: 1) it is used to generate bottom-up road transport emissions estimates, 2) their presence should be correlated with road transport emissions, and 3) it should be possible to obtain road network data globally either by using segmentation models applied to satellite data~\cite{van2018spacenet} or by sourcing it from governments or open-source databases like OSM~\cite{haklay2008openstreetmap}. 

To incorporate road network data, we use Rasterio~\cite{gillies_2019} to create GeoTIFFs co-registered with our Sentinel-2 swaths. These GeoTIFFs use road network data extracted from shapefiles provided by the Census TIGER system~\cite{marx1986tiger}. Several road categories are provided by Census TIGER, but we only use data with MAF/TIGER Feature Class Codes (MTFCC) S1100 (primary road), S1630 (ramp), S1200 (secondary road), and S1400 (local road). Primary road and ramp categories are merged to form the red channel of each GeoTIFF, while secondary roads and local roads form the green and blue channels, respectively. These road images match the CRS used by our Sentinel-2 swaths and are computationally inexpensive to dynamically sample during training as Sentinel-2 imagery is tiled. Sample input Sentinel-2 and road imagery can be seen in Figure \ref{fig:sample_results}.

\subsubsection{CO$_{2}$}

The benefits of using additional satellite and ground-based measurements of CO$_{2}$ concentrations were also examined in this work. The Orbiting Carbon Observatory-2 (OCO-2) satellite from NASA measures column-averaged CO$_{2}$ dry air mole fraction (X$_{\mathrm{CO}_{2}}$) in a sun-synchronous orbit with a 16 day revisit period~\cite{crisp2017orbit}. To reduce the level of effort in integrating this data into our model, and to address spatial, temporal, and data quality deficiencies in the OCO-2 Level 2 product~\cite{oco2level2product}, we focus on the interpolated OCO-2 Level 3 dataset from the University of Wollongong~\cite{zammit2018statistical}. To create the Level 3 product, fixed-rank Kriging is applied with a 16 day moving window in order to create global, daily, CO$_{2}$ concentration estimates.  Tabulated Level 3 products were first converted to global, annually-averaged GeoTIFFs for easier integration into the data processing pipeline. Due to the lower 1$^{\circ}$ x 1$^{\circ}$ ($\approx$100 x 100~km, depending on latitude) spatial resolution of the OCO-2 Level 3 product, typically only a single OCO-2 measurement covers an entire Sentinel-2/DARTE tile ($\approx$10 x 10~km). Thus, this value is used for all input image pixels in the form of an additional input channel.

Ground-based CO$_{2}$ measurements were also explored in order to determine whether measurements taken closer to road-level offered any measurable improvement in emissions estimation accuracy. NOAA's CarbonTracker project~\cite{peters2007atmospheric} offers monthly-averaged CO$_{2}$ mole fraction estimations at 1$^{\circ}$ x 1$^{\circ}$ spatial resolution over North America, and at varying levels of the atmosphere~\cite{carbontrackerCT2019Bproduct}. These concentrations are derived from their optimized surface flux product that incorporates 460 observation datasets from across the globe, recorded on the ground, in aircraft, and onboard ships. Concentrations are available at 26 geopotential heights, but we use values closest to the ground with an average altitude of 445~m. Similar to OCO-2, an annual average GeoTIFF is created and a single CarbonTracker value is used as an additional channel for each input tile.

\subsubsection{Population}

Population is likely to be correlated with road transport emissions, as vehicles are still primarily operated by people and there were 1.88 vehicles per household as of 2001-2007~\cite{fhwa2017tansport}. As such, we investigated incorporating population data into our model as an additional channel input. There are several sources of population data, with the most accurate source likely coming from a governmental censuses. However, our eventual goal for this work is to estimate road transport emissions globally, in a top-down manner, and without reliance on extensive data collection from local governments.

It is possible to achieve reasonable accuracy by estimating population from overhead imagery~\cite{hadzic2020estimating}. For this effort, we require annual (or sub-annual) gridded population estimates that can be paired with Sentinel-2 and DARTE data. LandScan estimates~\cite{Rose2018} provided by Oak Ridge National Laboratory meet this need. LandScan offers annual global population distribution GeoTIFFs from 2000 through 2019. For this effort, we use their 2017 product.

Each LandScan GeoTIFF contains ambient (24-hour average) population distribution estimates at a roughly 1~km$^2$ spatial resolution. We incorporate these population estimates by extracting tiles co-registered with our Sentinel-2 imagery and then concatenating them as an additional channel input to our model. Bilinear interpolation is used during both the CRS resampling process and for any upsampling needed to match the Sentinel-2 tile resolution. To ensure population consistency before and after this processing procedure, we normalize the resampled population tile to ensure that its total population matches the population total of the same region before resampling. Sample population data can be seen in Figure~\ref{fig:results_montage}.

\subsubsection{Dataset Partitioning}
The dataset we created for this work was constructed from data (e.g., satellite imagery, roads, etc.) depicting 3753 cities for training and 118 cities for validation, as visualized in Figure~\ref{fig:trainval_locations}. These cities are isolated to the conterminous United States (covering over 8~Mkm$^2$). 

Since the U.S. is vast and contains many sparsely populated regions, we construct our dataset by targeting regions that are most likely to have roads. We accomplish this by first creating a list of cities from the United States Cities Database~\cite{usaCities2021}, which is an aggregation of data collected by the U.S. Geological Survey and U.S. Census Bureau. Next, we designate cities for the validation split by identifying the nearest neighboring city to 108 airports across the U.S.~\cite{airportsUsaCities2021}, where at least one city is selected for each state. Since airports tend to be spatially dispersed across the U.S., the cities set aside for the validation set reflect many of the diverse geographies across the U.S. 

With validation cities identified, we construct the training split by removing all cities within a 120~km radius of the centerpoint of any validation city. This process prevents potential overlap in imagery across dataset splits. We then down-select from this filtered list, iterating over each state and omitting all cities not included in the 35 most populated cities, the 35 least populated cities, and 30 randomly-selected remaining cities. Should a state not have 100 cities in total, all cities are selected.

After partitioning the cities, we can begin to load tile tuples (i.e., co-registered data across our various sources) for training our models. However, as we load tiles we can encounter conditions that are not ideal for training machine learning models. If 50\% or more of a DARTE tile contains invalid data then we filter the tuple. Similarly, if more than 20\% of a Sentinel-2 tile is empty then we filter the tuple. We also filter out any tiles associated with data read errors.

After these filtering processes, we are left with 297,296 tiles (over 31~Mkm$^2$) for training and 45,224 tiles (over 4.7~Mkm$^2$) for validation. We further split our validation tiles into a random subset of 1000 tiles that are used to validate each epoch during model training and select optimal model weights, as well as 44,224 testing tiles that are used to generate the results in Tables~\ref{table:architecture_comparison}~and~\ref{table:input_comparison}.

\subsection{Architecture}

The two base architectures we used in our experiments were U-Net~\cite{ronneberger2015u} and MA-Net~\cite{fan2020ma}. U-Nets have been a popular choice for the winning solutions of several public challenges and datasets focused on per-pixel classification and regression in both medical imaging (where they were conceived) and satellite imagery \cite{bosch2019semantic,van2018spacenet,gupta2019creating}.
Both ResNet-34~\cite{he2016deep} and EfficientNet-B3~\cite{tan2019efficientnet} backbones were tested. Given that DARTE data has a lower resolution than Sentinel-2 imagery (1~km$^2$ vs 100~m$^2$), we also decided to modify the standard U-Net to perform reduced upsampling within the network. Given that inputs to the network had a 1024~x~1024 resolution, we only kept enough upsampling layers to result in an output size of 128~x~128, which is closer to the native DARTE resolution for a crop of the same geographic region. We refer to this architecture as the ``Reduced U-Net''. Given the shared success of U-Nets in both the medical imaging and satellite imagery domains, we also decided to test the MA-Net architecture due to its recent success in tumor segmentation applications. 

All model architecture and backbone implementations used in this work were built upon the implementations of \cite{Yakubovskiy:2019}. With each model using a ReLU activation for its final layer to regress positive per-pixel CO$_2$ values. Additionally, RAdam~\cite{liu2019variance} was used as the optimizer of choice for our experiments with a learning rate of 1e-3, betas of (0.9, 0.999), and eps of 1e-8. 

\begin{table}[t]
\centering
\begin{tabular}{l c c c}
\hline
Method & RMSLE & MAE & MAPE \\ [0.5ex]
\hline\hline
RN-34 U-Net & 0.661 & \textbf{38.9} & 50\% \\
EN-B3 U-Net & 0.71 & 51.2 & \textbf{47\%} \\ 
EN-B3 Reduced U-Net & 0.836 & 2669 & 214\% \\
EN-B3 MA-Net & \textbf{0.616} & 39.5 & 55\% \\ [0.5ex]
\hline
\end{tabular}
\caption{Comparison of results across varying neural network architectures trained on Sentinel-2 and road network data, including ResNet-34 (RN-34) and EfficientNet-B3 (EN-B3) backbone U-Nets, a Reduced U-Net architecture, and an MA-Net architecture. MAE is in units of kg CO$_{2}$ per 100~m$^{2}$.}
\label{table:architecture_comparison}
\end{table}

\subsection{Loss functions}
We experimented with different procedures for normalizing the DARTE data (e.g., 0-1 normalization, mean-std normalization, quantile transformation, K-bins discretization, etc.) along with different loss functions. We found the normalization approaches offered little benefit, while root mean square logarithmic error (RMSLE) and mean absolute percentage error (MAPE) loss functions were more promising. We believe this is likely due to large variations and outliers in road transport emissions across varying geographies, as illustrated in the histogram of DARTE emissions data in Figure~\ref{fig:darte_distribution}. For example, many rural areas have few roads and relatively little road transport emissions. Conversely, dense cities have high road densities and road transport emissions. Furthermore, city emissions can vary dramatically depending on factors such as public transit usage.

RMSLE is commonly used for regression tasks where the underlying ground truth data distribution is exponential or has many outliers, expressed as follows:

\begin{equation}
RMSLE=\sqrt{\frac{1}{n}\sum_{i=1}^{n} (log(P_{i}+1) - log(GT_{i}+1))^{2}}.\label{eq:rmsle}
\end{equation}

Our mean absolute percentage error (MAPE) metric is slightly modified from the typical MAPE equation to improve numerical stability as the ground truth ($GT$) approaches 0, which is defined as: 
\begin{equation}
    MAPE=\frac{100}{n}\sum_{i=1}^{n} \frac{|(GT_{i}+1) - (P_{i}+1)|}{GT_{i}+1}.\label{eq:mape}
\end{equation}

\begin{table}[t]
\centering
\begin{tabular}{l c c c}
\hline
Method & RMSLE & MAE & MAPE \\ [0.5ex]
\hline\hline
S2 & 1.03 & 55.9 & 65\% \\ 
R & 0.73 & 43.3 & 64\% \\ 
S2+R & \textbf{0.616} & \textbf{39.5} & 55\% \\
S2+OCO2 & 1.05 & 55.1 & 88\% \\
S2+R+LS & 0.739 & 49.9 & 47\% \\
S2+R+OCO2 & 0.709 & 49.3 & \textbf{46\%} \\
S2+R+OCO2+CT & 0.817 & 52.6 & \textbf{46\%} \\ [0.5ex]
\hline
\end{tabular}
\caption{Comparison of MA-Net results for models trained with varying inputs, including Sentinel-2 visual imagery (S2), road imagery (R), LandScan (LS) population estimates, Orbiting Carbon Observatory-2 (OCO2) Level 3 data, and CarbonTracker data (CT). MAE is in units of kg CO$_{2}$ per 100~m$^{2}$.}
\label{table:input_comparison}
\end{table}

In each equation, $n$ represents the number of pixels in each tile, $i$ represents the pixel index, $P$ represents the predicted emissions output from our model, and $GT$ represents the upsampled ground-truth emissions from DARTE.

In our experiments, we achieved the best performance by training using an RMSLE loss function. One added benefit to using RMSLE in the context of emissions estimation is that RMSLE is biased towards overestimating. In other words, underestimates incur a larger cost than overestimates. In the context of mitigating climate change, it is safer to overestimate emissions and take more drastic action than necessary as opposed to underestimating emissions and not taking sufficient action.

\section{Results \& Discussion}
\label{ch:results}

\begin{figure*}[t]
    \centering
    \includegraphics[width=\textwidth]{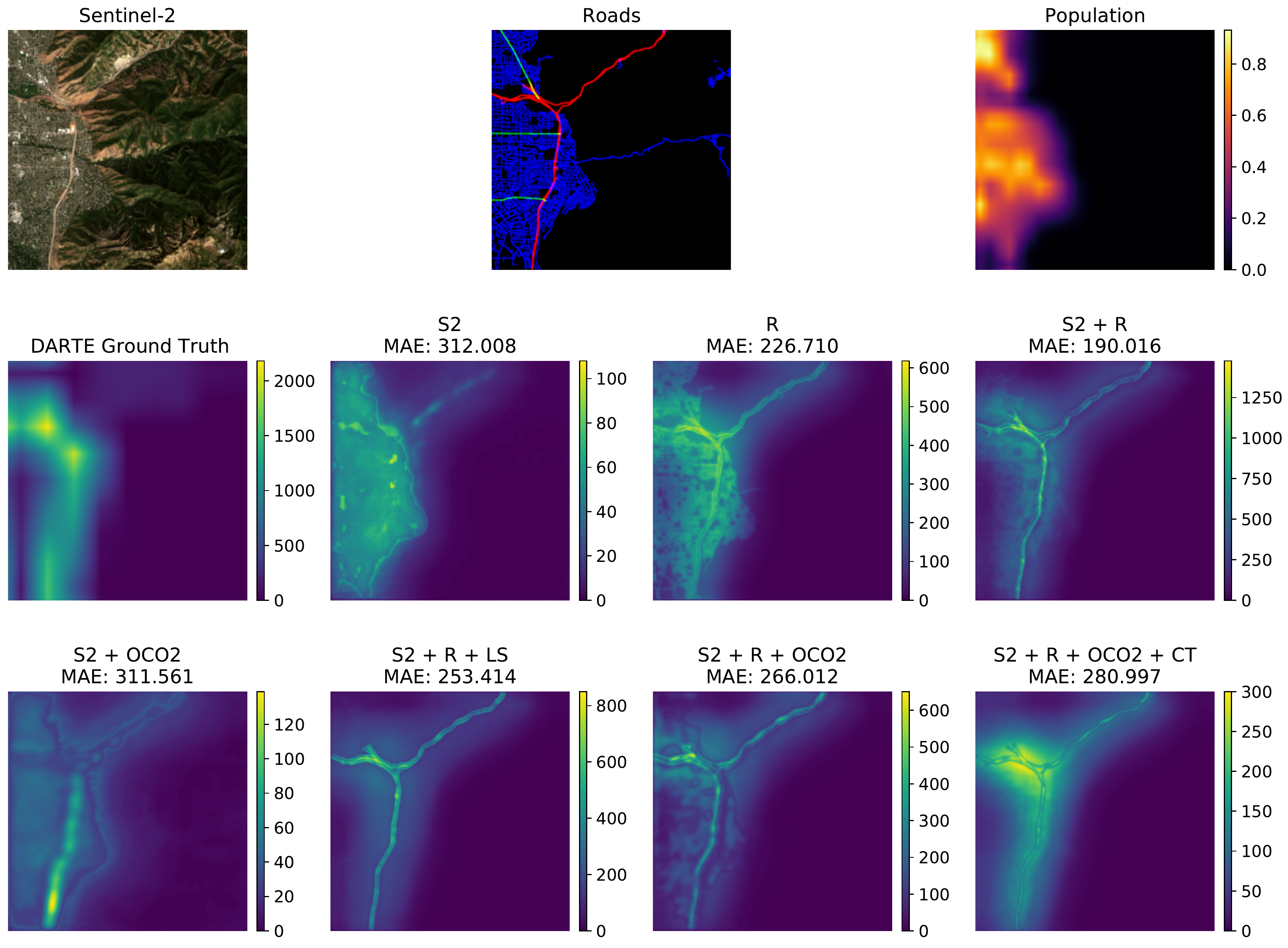}
    \caption{Example results from the seven models listed in Table~\ref{table:input_comparison}. Emissions colorbars are in units of kg CO$_{2}$ per 100~m$^{2}$, and LandScan population is in units of people per 100~m$^{2}$.}
    \label{fig:results_montage}
\end{figure*}

We evaluate multiple segmentation architectures and backbones against our reduced Sentinel-2 and roads dataset to obtain the results shown in Table~\ref{table:architecture_comparison}. An MA-Net architecture with EfficientNet-B3 backbone achieves the best results with an RMSLE of 0.616, while ResNet-34 and EfficientNet-B3 backboned U-Nets achieve slightly better MAE and MAPE. 

We believe one of the biggest challenges associated with training these models comes from the spatial resolution mismatch between our predictions and the ground truth. Figures~\ref{fig:sample_results}~and~\ref{fig:results_montage} qualitatively illustrate that our EfficientNet-B3 backbone MA-Net model is capable of learning that road transport emissions are highest on roads. Taking notice of the the bright yellow region near the bottom-center of the ``Predicted Emissions'' image. Instead of reproducing the blurry yellow region of emissions located at the same region in the ``Ground Truth Emissions'' image, this model can predict large emissions with fine-grained structure matching the roads present in the scene. However, by calculating error in a pixel-wise fashion, the model is penalized for learning this fine-grained structure. In other words, if the model correctly estimates low emissions from nearby farmland, it will be penalized due to the fact that our ground truth contains large emissions values in that area.

We believe that one potential solution for this spatial resolution mismatch is to compare model predictions at the same resolution as the ground truth. Our Reduced U-Net architecture represents a first attempt to accomplish this. Rather than outputting 1024~x~1024 pixel predictions that matches the spatial resolution of its input, our Reduced U-Net outputs a 128~x~128 prediction to more closely match the spatial resolution of our ground truth. Unfortunately, as can be seen in Table~\ref{table:architecture_comparison}, this approach does not perform well, resulting in significantly higher RMSLE, MAE, and MAPE. We hypothesize that the poor performance of this model is due, at least in part, to the reduction of skip connections, especially from high-resolution layers, which may be necessary to exploit the small and subtle visual features (e.g., roads, buildings, etc.) in our imagery. We suspect that simply adding additional downsampling layers after the segmentation architecture may perform better, but leave this experiment for future work.

In Section~\ref{ch:method}, we outline several data sources that we believe may improve our model's accuracy and ability to generalize. As shown in Table~\ref{table:input_comparison}, we trained multiple EfficientNet-B3 MA-Net models on varying permutations of these input data sources. Sample results from these models can be seen in Figure~\ref{fig:results_montage}. The best performing model was trained using Sentinel-2 imagery and road network data. This model outperformed both the Sentinel-2 only and Roads-only models, as we expected. However, results from models trained with additional inputs from OCO-2, LandScan, and CarbonTracker all underperform compared to the model using only Sentinel-2 and road data. We hypothesize that this may also be related to the spatial resolution mismatch, except in this instance not only is there a mismatch between the inputs and our ground truth but there is a mismatch between different input data sources. Road data is actually rasterized to match the spatial resolution of the input Sentinel-2 data, whereas all other input data sources have different underlying resolutions. It is possible that this spatial resolution mismatch between input data sources introduces an additional challenge by implying certain image regions are more similar than the higher-resolution data indicates. Late fusion or other techniques that account for this mismatch may need to be investigated to improve exploitation of these inputs.

Another aspect to consider in reviewing our model results is the overall difficulty of our dataset. The included cities and neighboring regions contain diverse geographies and 5 of the 6 Köppen-Geiger climate zone types, which can be visually representative of many regions across the globe. This visual diversity adds challenge, but also helps inform the ability of the model to generalize globally.

Additionally, as shown in Figure~\ref{fig:trainval_locations}, our validation split sequesters many of the largest east and west-coast cities from training. Many of these cities represent significant outliers, both visually and in terms of their emissions, compared to most other U.S. cities. However, the challenge of this split may also offer insight towards global generalizability. Road transport emissions estimation methods must learn to generalize not only across diverse visual backgrounds, as discussed previously, but also across diverse architectures, population densities, and transport systems located across small, medium, and large cities.

A similar point can also be made concerning our data sampling procedure. The population density or urbanization distribution of tiles in our dataset is likely skewed, matching the real world distribution. However, for the purposes of training a model that can estimate urban road transport emissions as accurately as rural road transport emissions, it may be necessary to oversample the less-frequently-occuring urban areas, especially as estimating urban road transport emissions is likely to be a more challenging task.

\section{Conclusion}
\label{ch:conclusion}

We present the first published automated approach for estimating road transport sector emissions in a top-down fashion using visual satellite imagery. Using primarily Sentinel-2 imagery and road network data over the conterminous United States, we train an MA-Net segmentation model to regress road transport emissions on a pixel-by-pixel basis using an RMSLE loss. Our model achieves an MAE of 39.5~kg~CO$_{2}$ per 100~m$^{2}$.

Significant challenges remain to operationalize this technology. Model accuracy must be improved and estimation uncertainty quantified to provide actionable information for regional governments and municipalities. The limited availability of accurate and global road transportation emissions data must be overcome, including concerns of model transfer and regional bias. It remains to be seen if changes in government policy and human behavior over annual timescales will be captured by our models, although this hypothesis will be testable as data from 2020 emerges. Despite these challenges, we believe this work represents a critical step towards building scalable, global, near-real-time road transportation emissions inventories that can provide independent and objective feedback as the global community tackles climate change.

\section{Acknowledgements}
This research was conducted as part of the Climate TRACE initiative to track global GHG emissions and make the data publicly available. Financial support was provided by a grant from Generation Investment Management. We would also like to thank Gabriela Volpato and Aaron Davitt from WattTime, as well as Mathieu Carlier, B\'en\'edicte De Gelder, and Alain Reti\`ere from Everimpact for their support identifying data sources and discussing methodology.

{\small
\bibliographystyle{ieee_fullname}
\bibliography{egbib}
}


\end{document}